\documentclass[sigconf]{acmart}
\renewcommand\footnotetextcopyrightpermission[1]{}
\usepackage{booktabs} 
\graphicspath{ {images/} }

\usepackage{tabularx}
\usepackage{subfig}
\usepackage{xcolor}
\usepackage{standalone}
\usepackage{tikz}
\usetikzlibrary{snakes,arrows,shapes}
\usepackage{pgfplots}

\newcommand\cubesize[1]{$(#1\times #1\times #1)$}

\setcopyright{rightsretained}

\acmConference[CIKM'17]{Conference on Information and Knowledge Management}{November 2017}{Singapore}
\acmYear{2017}
\copyrightyear{2017}

\begin{document}
\title{Deep Learning for Lung Cancer Detection: \\ Tackling the Kaggle Data Science Bowl 2017 Challenge}

\author{Kingsley Kuan}
\email{kingsley.kuan@gmail.com}
\authornote{Authors contributed equally to this work.}
\affiliation{\institution{Institute for Infocomm Research}}

\author{Mathieu Ravaut\footnotemark[1]}
\email{mathieu.ravaut@student.ecp.fr}
\affiliation{\institution{Institute for Infocomm Research}}
\affiliation{\institution{CentraleSup\'elec}}

\author{Gaurav Manek}
\email{manekgm@i2r.a-star.edu.sg}
\affiliation{\institution{Institute for Infocomm Research}}

\author{Huiling Chen}
\email{chenhl@i2r.a-star.edu.sg}
\affiliation{\institution{Institute for Infocomm Research}}

\author{Jie Lin}
\email{lin-j@i2r.a-star.edu.sg}
\affiliation{\institution{Institute for Infocomm Research}}

\author{Babar Nazir}
\email{babar.nazir@singhealth.com.sg}
\affiliation{\institution{National Cancer Centre Singapore}}

\author{Cen Chen}
\email{chencen@hnu.edu.cn}
\affiliation{\institution{Hunan University}}

\author{Tse Chiang Howe}
\email{howetc@live.com}
\affiliation{\institution{Chesed Radiology}}

\author{Zeng Zeng}
\email{zengz@i2r.a-star.edu.sg}
\authornote{Corresponding authors.}
\affiliation{\institution{Institute for Infocomm Research}}

\author{Vijay Chandrasekhar\footnotemark[2]}
\email{vijay@i2r.a-star.edu.sg}
\affiliation{\institution{Institute for Infocomm Research}}
\affiliation{\institution{Nanyang Technological University}}

\renewcommand{\shortauthors}{K. Kuan, M. Ravaut, et al.}

\begin{abstract}
We present a deep learning framework for computer-aided lung cancer diagnosis. Our multi-stage framework detects nodules in 3D lung CAT scans, determines if each nodule is malignant, and finally assigns a cancer probability based on these results. We discuss the challenges and advantages of our framework. In the Kaggle Data Science Bowl 2017, our framework ranked $41^\text{st}$ out of 1972 teams.
\end{abstract}

%
%



\maketitle

\section{Introduction}
\label{sec:intro}

Cancer is one of the leading causes of death worldwide, with lung cancer being among the leading cause of cancer related death. In 2012, it was estimated that 1.6 million deaths were caused by lung cancer, while an additional 1.8 million new cases were diagnosed \cite{CAAC:CAAC21262}.

Screening for lung cancer is crucial in the early diagnosis and treatment of patients, with better screening techniques leading to improved patient outcome. The National Lung Screening Trial found that screening with the use of low-dose helical computed tomography (CT) reduced mortality rates by 20\% compared to single view radiography in high-risk demographics \cite{doi:10.1056/NEJMoa1102873}. However, screening for lung cancer is prone to false positives, increasing costs through unnecessary treatment and causing unnecessary stress for patients \cite{doi:10.1001/jamainternmed.2013.12738}. Computer-aided diagnosis of lung cancer offers increased coverage in early cancer screening and a reduced false positive rate in diagnosis.

The Kaggle Data Science Bowl 2017 (KDSB17) challenge was held from January to April 2017 with the goal of creating an automated solution to the problem of lung cancer diagnosis from CT scan images \cite{kaggle_2017}. In this work, we present our solution to this challenge, which uses 3D deep convolutional neural networks for automated diagnosis.

\subsection{Related Work}

Computer-aided diagnosis (CAD) is able to assist doctors in understanding medical images, allowing for cancer diagnosis with greater sensitivity and specificity, which is critical for patients. \citet{histosurvey} survey CAD pipelines, separating them into preprocessing, feature extraction, selection, and finally classification. They further document the use of logistic regression, decision trees, k-nearest neighbour, and neural networks in existing approaches.

\citet{prostate} use an SVM over MRI scan texture features to detect prostate cancer in patients. The winners of the Camelyon16 challenge \cite{Camelyon16}, for example, detect breast cancer from images of lymph nodes.

Deep convolutional neural networks (CNN) have proven to perform well in image classification \cite{NIPS2012_4824, Simonyan14c, he2016deep}, object detection \cite{renNIPS15fasterrcnn}, and other visual tasks. They have found great success in medical imaging applications \cite{KayalibayJS17}, and are for example able to detect skin cancer metastases \cite{googlemetastasis}, achieving substantially better sensitivity performance than human pathologists. These methods all operate on two-dimensional images, typically a cross-sectional image of the affected body part.

In comparison, CT image scans are three-dimensional volumes and are usually anisotropic. Deep networks have also been shown to perform well in 3D segmentation \cite{vnet}, and have been successfully adapted from 2D to 3D \cite{unet3d,voxresnet}. \citet{prostatesegm} have demonstrated a 3D deep learning framework to perform automatic prostate segmentation. \citet{nodulesAE} perform the segmentation and classification of lung cancer nodules separately.

The LUNA16 challenge \cite{DBLP:journals/corr/SetioTBBBC0DFGG16} had two tasks: detecting pulmonary nodules using CT scans, and reducing the false positive rate from identifying these nodules. The former was solved by \citet{znet} using UNet \cite{unet} on stacks of 3 consecutive horizontal lung slices; and the latter was won by \citet{fprnodule} by applying multi-contextual 3D CNNs.

The top two teams in the Kaggle Data Science Bowl 2017 have published their solutions to the challenge \cite{grt123, juliandewit}. Both teams proceed to an intermediate step before giving patients a cancer probability. After having identified regions of possible abnormalities (nodules), the team placing second, \citet{juliandewit}, uses 17 different 3D CNNs to extract medically relevant features about nodules. Then, they aggregate the predictions of these nodules attributes into a patient-level descriptor. The team placing first \cite{grt123} detects nodules via a 3D CNN, then uses the highest confidence detections as well as manual nodule labelling to predict cancer via a simple classifier.

\subsection{Key Challenges}

One key characteristic of lung cancer is the presence of \emph{pulmonary nodules}, solid clumps of tissue that appear in and around the lungs \cite{noduleinfo}. These nodules are visible in CT scan images and can be \emph{malignant} (cancerous) in nature, or \emph{benign} (not cancerous). In cancer screening, radiologists and oncologists examine CT scans of the lung volume to identify nodules and recommend further action: monitoring, blood tests, biopsy, etc. Specifically, lung cancer is screened through the presence of nodules \cite{noduleinfo, cancerus}.

To build a system for computer-aided diagnosis (CAD) of lung cancer, we investigate the following approaches:

\begin{enumerate}
  \item A \textbf{single-stage} network that automatically learns cancer features and produces a cancer probability directly from CT scan images of patients.
  \item A \textbf{multi-stage} framework that first localizes nodules, classifies the malignancy of each one, and finally produces a cancer probability of patients.
\end{enumerate}

In our initial experiments however, the single-stage network, implemented as a 3D CNN, fails to converge across a wide set of hyperparameters, performing only slightly better than random chance.

\begin{figure}[h]
\centering
\begin{tikzpicture}
\begin{axis}
[
xlabel=Steps,
ylabel=Loss,
xlabel near ticks,
ylabel near ticks,
no markers
]
\addplot
table [col sep=comma, x=Step, y=Value]
{figures/classifier_loss.csv};
\end{axis}
\end{tikzpicture}

\begin{tikzpicture}
\begin{axis}
[
xlabel=Steps,
ylabel=Loss,
xlabel near ticks,
ylabel near ticks,
no markers
]
\addplot
table [col sep=comma, x=Step, y=Value]
{figures/cancer_detector_loss.csv};
\end{axis}
\end{tikzpicture}

\caption{Top: Training loss of \textit{single-stage} network, which fails to converge. \\ Bottom: Training loss of malignancy detector in the \textit{multi-stage} framework, which easily converges.}
\label{fig:training_loss}
\end{figure}
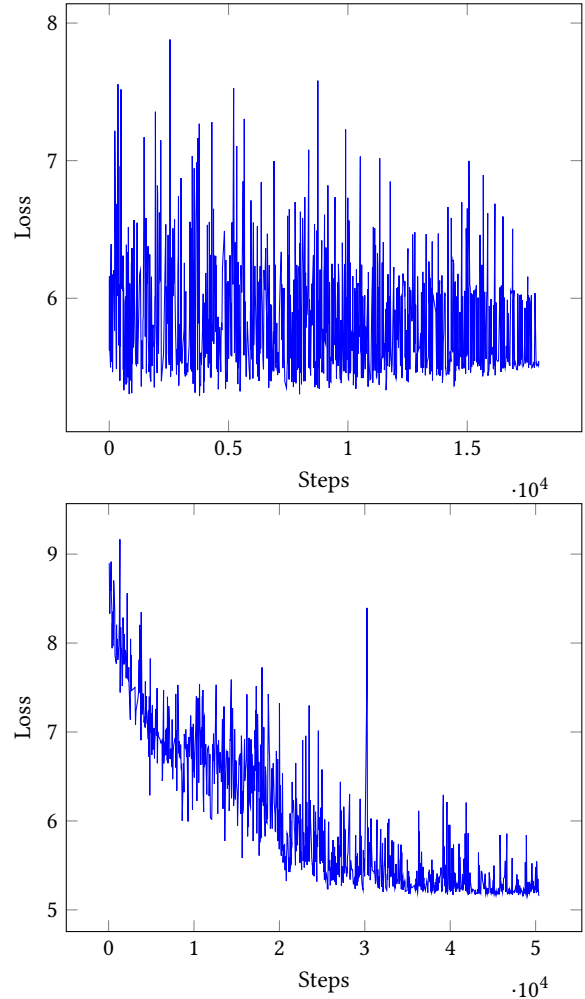

Factoring the problem into multiple stages on the other hand significantly improves convergence. Even when the single-stage network fails to converge in training, each stage of our pipeline can be easily trained to convergence, as illustrated in Figure~\ref{fig:training_loss}. Our framework as detailed in this work therefore focuses on a multi-stage pipeline, focusing on detection and classification of pulmonary nodules.

This approach presents the following problems: the shape and size of nodules vary, and benign nodules can look very similar to malignant ones. Furthermore, the presence of blood vessels in the lung makes distinguishing nodules a challenging task, especially on 2D image slices. This makes the task more suitable for 3D CNNs which are better able to identify nodules based on their structure in 3D space.

\section{Technical Approach}
\subsection{Data}

While the Kaggle Data Science Bowl 2017 (KDSB17) dataset provides CT scan images of patients, as well as their cancer status, it does not provide the locations or sizes of pulmonary nodules within the lung. Therefore, in order to train our multi-stage framework, we utilise an additional dataset, the Lung Nodule Analysis 2016 (LUNA16) dataset, which provides nodule annotations. This presents its own problems however, as this dataset does not contain the cancer status of patients. We thus utilise both datasets to train our framework in two stages.

\subsubsection{LUNA16}

The Lung Nodule Analysis 2016 (LUNA16) dataset is a collection of 888 axial CT scans of patient chest cavities taken from the LIDC/IDRI database\cite{armato2011lung}, where only scans with a slice thickness smaller than 2.5 mm are included. In each scan, the location and size of nodules are agreed upon by at least 3 radiologists. There is no information regarding the malignancy or benignity of each nodule or the cancer status of the associated patient.

\begin{figure}[h]
\includegraphics[width=\columnwidth,trim={0cm 1.5cm 0cm 1.5cm},clip]{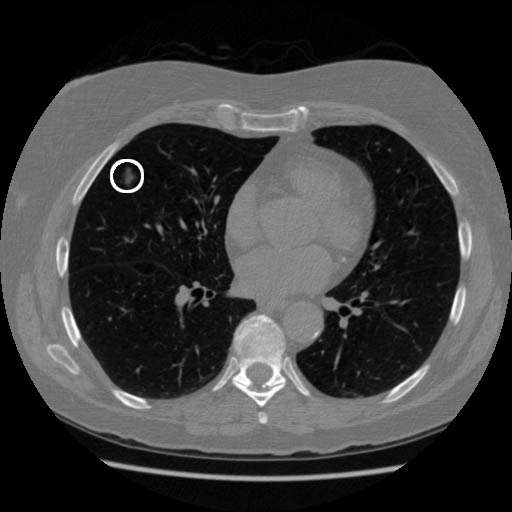}
\caption{A slice displaying a cross section of a patient chest cavity from the LUNA16 dataset, with a nodule annotated.}
\label{fig:lunapic}
\end{figure}

In total, 1186 nodules are annotated across 601 patients. We use 542 patients as a training set and the remaining 59 as a validation set. A slice from one patient, with a single nodule location shown, can be seen in Figure~\ref{fig:lunapic}.

\subsubsection{Kaggle Data Science Bowl 2017}

The Kaggle Data Science Bowl 2017 (KDSB17) dataset is comprised of 2101 axial CT scans of patient chest cavities. Of the 2101, 1595 were initially released in stage 1 of the challenge, with 1397 belonging to the training set and 198 belonging to the testing set. The remaining 506 were released in stage 2 as a final testing set.

\begin{figure}[h]
\includegraphics[width=\columnwidth,trim={1cm 3.5cm 2cm 3cm},clip]{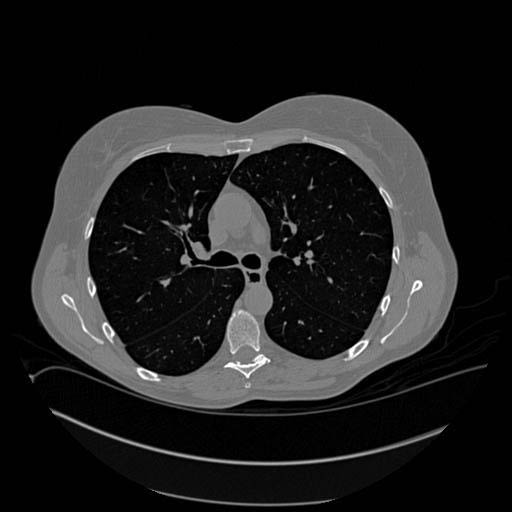}
\caption{A slice displaying a cross section of a patient chest cavity from the Kaggle Data Science Bowl 2017 dataset. No annotations of nodule locations or sizes are provided.}
\label{fig:kagglepic}
\end{figure}

Each CT scan was labelled as `with cancer' if the associated patient was diagnosed with cancer within 1 year of the scan, and `without cancer' otherwise. Crucially, the location or size of nodules are not labelled. Figure~\ref{fig:kagglepic} contains a sample slice from this dataset.

\subsubsection{Preprocessing}
Each scan is comprised of multiple 2D axial scans taken in sequence with pixel values in the range (-1024, 3071), corresponding to Hounsfield radiodensity units. The number of slices, slice thickness, and scale vary between scans.

We normalize pixel values to the (0, 1) range and stack the 2D slices in sequence to produce a 3D volume. The entire 3D volume is scaled and padded, while maintaining the true aspect ratio using the embedded scale information, into a \cubesize{512} volume. Due to the increased GPU memory usage involved with volumetric data, we then separate this volume into overlapping \cubesize{128} crops with a 64 voxel stride, to be processed by our pipeline in parallel.

\subsection{Architecture}

Our framework is divided into four separate neural networks. They are the:
\begin{enumerate}
  \item \textbf{nodule detector}, which accepts a normalized 3D volume crop from a CT scan and identifies areas that contain nodules;
  \item \textbf{malignancy detector}, which operates similarly to the nodule detector, but further classifies nodules as benign (non-cancerous) or malignant (cancerous);
  \item \textbf{nodule classifier}, which accepts individual nodule volumes and similarly classifies them as benign or malignant; and--
  \item \textbf{patient classifier}, which accepts features from the malignancy detector and nodule classifier and yields the probability of the patient having cancer.
\end{enumerate}

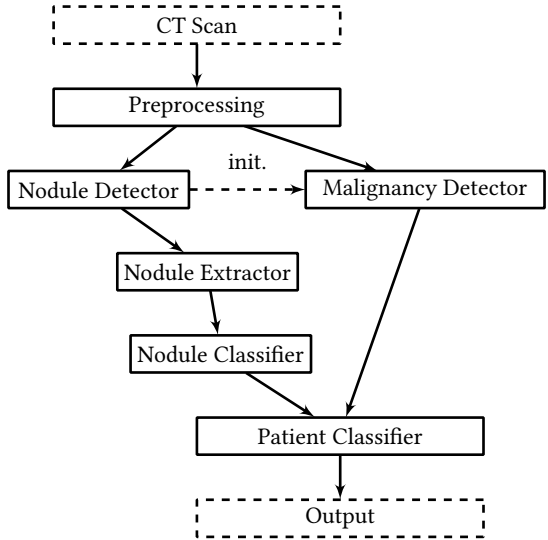
\begin{figure}[h]
\pagestyle{empty}

\enlargethispage{100cm}
\begin{tikzpicture}[>=latex',line join=bevel,]
  \pgfsetlinewidth{1bp}
\begin{scope}
  \pgfsetstrokecolor{black}
  \definecolor{strokecol}{rgb}{1.0,1.0,1.0};
  \pgfsetstrokecolor{strokecol}
  \definecolor{fillcol}{rgb}{1.0,1.0,1.0};
  \pgfsetfillcolor{fillcol}
  \filldraw (0.0bp,0.0bp) -- (0.0bp,200.0bp) -- (202.0bp,200.0bp) -- (202.0bp,0.0bp) -- cycle;
\end{scope}
  \pgfsetcolor{black}
  \draw [->,dashed] (68.262bp,131.0bp) .. controls (78.603bp,131.0bp) and (90.231bp,131.0bp)  .. (111.88bp,131.0bp);
  \definecolor{strokecol}{rgb}{0.0,0.0,0.0};
  \pgfsetstrokecolor{strokecol}
  \draw (90.0bp,141.5bp) node {init.};
  \draw [->] (89.543bp,61.85bp) .. controls (94.555bp,58.62bp) and (100.87bp,54.552bp)  .. (115.47bp,45.141bp);
  \draw [->] (88.819bp,154.99bp) .. controls (100.58bp,151.03bp) and (116.22bp,145.75bp)  .. (139.2bp,138.0bp);
  \draw [->] (154.8bp,123.75bp) .. controls (149.84bp,109.65bp) and (137.68bp,75.053bp)  .. (127.13bp,45.059bp);
  \draw [->] (63.153bp,154.85bp) .. controls (59.327bp,151.85bp) and (54.578bp,148.13bp)  .. (41.835bp,138.14bp);
  \draw [->] (42.695bp,123.85bp) .. controls (47.152bp,120.7bp) and (52.739bp,116.75bp)  .. (66.317bp,107.14bp);
  \draw [->] (125.0bp,30.85bp) .. controls (125.0bp,28.851bp) and (125.0bp,26.53bp)  .. (125.0bp,14.141bp);
  \draw [->] (71.0bp,185.85bp) .. controls (71.0bp,183.85bp) and (71.0bp,181.53bp)  .. (71.0bp,169.14bp);
  \draw [->] (76.06bp,92.85bp) .. controls (76.405bp,90.851bp) and (76.805bp,88.53bp)  .. (78.941bp,76.141bp);
\begin{scope}
  \definecolor{strokecol}{rgb}{0.0,0.0,0.0};
  \pgfsetstrokecolor{strokecol}
  \definecolor{fillcol}{rgb}{1.0,1.0,1.0};
  \pgfsetfillcolor{fillcol}
  \filldraw (125.0bp,169.0bp) -- (17.0bp,169.0bp) -- (17.0bp,155.0bp) -- (125.0bp,155.0bp) -- cycle;
  \draw (71.0bp,162.0bp) node {Preprocessing};
\end{scope}
\begin{scope}
  \pgfsetdash{{3pt}{3pt}}{0pt}
  \definecolor{strokecol}{rgb}{0.0,0.0,0.0};
  \pgfsetstrokecolor{strokecol}
  \draw [dashed] (179.0bp,14.0bp) -- (71.0bp,14.0bp) -- (71.0bp,0.0bp) -- (179.0bp,0.0bp) -- cycle;
  \draw (125.0bp,7.0bp) node {Output};
\end{scope}
\begin{scope}
  \definecolor{strokecol}{rgb}{0.0,0.0,0.0};
  \pgfsetstrokecolor{strokecol}
  \definecolor{fillcol}{rgb}{1.0,1.0,1.0};
  \pgfsetfillcolor{fillcol}
  \filldraw (179.0bp,45.0bp) -- (71.0bp,45.0bp) -- (71.0bp,31.0bp) -- (179.0bp,31.0bp) -- cycle;
  \draw (125.0bp,38.0bp) node {Patient Classifier};
\end{scope}
\begin{scope}
  \definecolor{strokecol}{rgb}{0.0,0.0,0.0};
  \pgfsetstrokecolor{strokecol}
  \definecolor{fillcol}{rgb}{1.0,1.0,1.0};
  \pgfsetfillcolor{fillcol}
  \filldraw (114.0bp,76.0bp) -- (46.0bp,76.0bp) -- (46.0bp,62.0bp) -- (114.0bp,62.0bp) -- cycle;
  \draw (80.0bp,69.0bp) node {Nodule Classifier};
\end{scope}
\begin{scope}
  \pgfsetdash{{3pt}{3pt}}{0pt}
  \definecolor{strokecol}{rgb}{0.0,0.0,0.0};
  \pgfsetstrokecolor{strokecol}
  \draw [dashed] (125.0bp,200.0bp) -- (17.0bp,200.0bp) -- (17.0bp,186.0bp) -- (125.0bp,186.0bp) -- cycle;
  \draw (71.0bp,193.0bp) node {CT Scan};
\end{scope}
\begin{scope}
  \definecolor{strokecol}{rgb}{0.0,0.0,0.0};
  \pgfsetstrokecolor{strokecol}
  \definecolor{fillcol}{rgb}{1.0,1.0,1.0};
  \pgfsetfillcolor{fillcol}
  \filldraw (109.0bp,107.0bp) -- (41.0bp,107.0bp) -- (41.0bp,93.0bp) -- (109.0bp,93.0bp) -- cycle;
  \draw (75.0bp,100.0bp) node {Nodule Extractor};
\end{scope}
\begin{scope}
  \definecolor{strokecol}{rgb}{0.0,0.0,0.0};
  \pgfsetstrokecolor{strokecol}
  \definecolor{fillcol}{rgb}{1.0,1.0,1.0};
  \pgfsetfillcolor{fillcol}
  \filldraw (202.0bp,138.0bp) -- (112.0bp,138.0bp) -- (112.0bp,124.0bp) -- (202.0bp,124.0bp) -- cycle;
  \draw (157.0bp,131.0bp) node {Malignancy Detector};
\end{scope}
\begin{scope}
  \definecolor{strokecol}{rgb}{0.0,0.0,0.0};
  \pgfsetstrokecolor{strokecol}
  \definecolor{fillcol}{rgb}{1.0,1.0,1.0};
  \pgfsetfillcolor{fillcol}
  \filldraw (68.0bp,138.0bp) -- (0.0bp,138.0bp) -- (0.0bp,124.0bp) -- (68.0bp,124.0bp) -- cycle;
  \draw (34.0bp,131.0bp) node {Nodule Detector};
\end{scope}
\end{tikzpicture}

%
\caption{Our framework design. The nodule detector is used to detect and extract nodules for the nodule classifier, and is also used to initialise the weights of the malignancy detector. The patient classifier aggregates results from both networks and produces the probability of the patient having lung cancer.}
\label{fig:framework}
\end{figure}

The nodule detector is used to detect regions that contain nodules. The malignancy detector provides a class probability map over grid cells of each cell containing benign, malignant, or no nodules. Separate code extracts and preprocesses nodule volumes and runs the classifier on each, yielding the probability of malignancy for each nodule. Finally, the patient classifier pools features from the classifier and features from the malignancy detector; producing the probability of the patient having lung cancer. Figure~\ref{fig:framework} graphically shows the structure of our pipeline.

\subsubsection{Nodule Detectors}

Common frameworks for object detection (such as Faster RCNN \cite{renNIPS15fasterrcnn}) produce precise bounding boxes around objects of interest. As our task does not require perfect localization, we instead divide the search space into a uniform grid and perform detection in each grid cell. This was inspired by the class probability map of the YOLO network \cite{redmon2016you}.

We base our architecture on the pre-activation version of ResNet-101 \cite{he2016deep,he2016identity}, which uses fewer parameters than other state-of-the-art networks while achieving comparably high accuracy on visual tasks. Our modified ResNet is described in Table~\ref{tab:nodule_detector}. Notably, we use 3D convolutions and pooling, and replace the global average pooling operation with a \cubesize{1} convolution. Additionally, we substitute rectified linear units (ReLU) with Leaky ReLU units ($\alpha = 0.1$), to improve convergence.

\begin{table}[h]
\centering
\begin{tabular}{lccr}
\toprule
\textbf{Layers} & \textbf{Size, Filters} & \textbf{Stride} & \textbf{Output Size}
\\ \midrule
Convolution 3D & \(7 \times 7 \times 7\), 64 & 2 & \(64 \times 64 \times 64\)
\\
Max Pool 3D & \(3 \times 3 \times 3\) & 2 & \(32 \times 32 \times 32\)
\\ \midrule \addlinespace[0.5em]
Block 1 & \(\begin{bmatrix}
1 \times 1 \times 1, 64 \\
3 \times 3 \times 3, 64 \\
1 \times 1 \times 1, 256 \\
\end{bmatrix} \times 3\) & 2 & \(16 \times 16 \times 16\)
\\ \addlinespace[0.5em]
Block 2 & \(\begin{bmatrix}
1 \times 1 \times 1, 128 \\
3 \times 3 \times 3, 128 \\
1 \times 1 \times 1, 512 \\
\end{bmatrix} \times 4\) & 2 & \(8 \times 8 \times 8\)
\\ \addlinespace[0.5em]
Block 3 & \(\begin{bmatrix}
1 \times 1 \times 1, 256 \\
3 \times 3 \times 3, 256 \\
1 \times 1 \times 1, 1024 \\
\end{bmatrix} \times 23\) & 1 & \(8 \times 8 \times 8\)
\\ \addlinespace[0.5em]
Block 4 & \(\begin{bmatrix}
1 \times 1 \times 1, 512 \\
3 \times 3 \times 3, 512 \\
1 \times 1 \times 1, 2048 \\
\end{bmatrix} \times 3\) & 1 & \(8 \times 8 \times 8\)
\\ \addlinespace[0.5em] \midrule
Convolution 3D & \(1 \times 1 \times 1\), $c$ & 1 & \(8 \times 8 \times 8\)
\\
Softmax &  &  & \(8 \times 8 \times 8\)
\\ \bottomrule
\end{tabular}
\caption{ResNet-101, modified for nodule detection (with $c=2$) and malignancy detection (with $c=3$). Notably, we use 3D operations, replace the global average pooling operation with a \cubesize{1} convolution, and narrow the receptive field of each output node to \cubesize{16} voxels.}
\label{tab:nodule_detector}
\end{table}

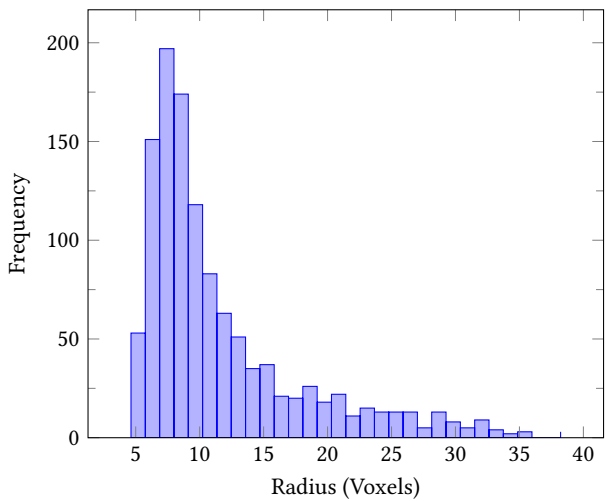
\begin{figure}[h]
\centering
\begin{tikzpicture}
\begin{axis}
[
xlabel=Radius (Voxels),
ylabel=Frequency,
xlabel near ticks,
ylabel near ticks,
xtick distance=5,
minor y tick num = 1,
ymin=0,
no markers,
area style
]
\addplot+[ybar interval] plot coordinates
{
(4.6314519246419268, 53)
(5.751041535408266, 151)
(6.8706311461746044, 197)
(7.9902207569409436, 174)
(9.1098103677072828, 118)
(10.229399978473623, 83)
(11.348989589239959, 63)
(12.4685792000063, 51)
(13.58816881077264, 35)
(14.707758421538976, 37)
(15.827348032305316, 21)
(16.946937643071657, 20)
(18.066527253837993, 26)
(19.186116864604333, 18)
(20.305706475370673, 22)
(21.42529608613701, 11)
(22.54488569690335, 15)
(23.66447530766969, 13)
(24.784064918436027, 13)
(25.903654529202367, 13)
(27.023244139968707, 5)
(28.142833750735043, 13)
(29.262423361501384, 8)
(30.382012972267724, 5)
(31.50160258303406, 9)
(32.621192193800404, 4)
(33.740781804566744, 2)
(34.860371415333084, 3)
(35.979961026099417, 0)
(37.099550636865757, 0)
(38.219140247632097, 3)
(38.219140247632097, 0)
};
\end{axis}
\end{tikzpicture}
\caption{Distribution of nodule radii in the LUNA16 dataset after preprocessing.}
\label{fig:nodule_sizes}
\end{figure}

We interpret the output as a class probability map of nodules occurring inside the corresponding receptive field. Our modifications also narrow the receptive field of each output node to \cubesize{16} voxels to better localize nodules as illustrated in Figure~\ref{fig:nodule_sizes}, which shows the distribution of nodule radii in the LUNA16 dataset.

\begin{figure}[h]
\includegraphics[width=\columnwidth]{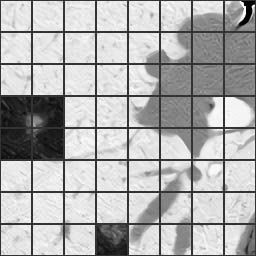} \\
\caption{Output class probability map of the nodule detector with \(c = 2\).}
\label{fig:detection_map}
\end{figure}

A \cubesize{128} input volume produces an \cubesize{8} output class probability map. The nodule detector provides a distribution over two classes \{`has-nodule', `no-nodule'\}, while the malignancy detector provides a distribution over three \{`malignant', `benign', `no-nodule'\}. An example of the output map can be seen in Figure~\ref{fig:detection_map}.

The output of the malignant detector is provided directly to the patient-level classifier. It acts as a global feature, providing information on the distribution of malignant nodules through the entire volume without providing specific information about any nodule.

As nodules are sparsely distributed through the scan, we expect there to be a strong class imbalance. We address this by weighting our cross-entropy function during training. In the nodule detector, we balance the loss by calculating a weight per-batch and apply it to the weaker class as in Equation~\ref{eq:bal_loss}.

\begin{align}
\textsc{Loss}(p, q) &= -\frac{1}{|C|} \sum_{c \in C} w(c) \cdot p(c) \log q(c) \label{eq:bal_loss}
\\ \text{where}~ & p ~\text{is the predicted distribution,}  \nonumber
\\ & q ~\text{is the true distribution,} \nonumber
\\ & C ~\text{is} ~\{\text{`no-nodule', `has-nodule'}\} \nonumber
\\ & f_c ~\text{is the frequency of class $c$ in the mini-batch}\text{, and--}  \nonumber
\\ & w(c) = \left\{\begin{array}{lr}
   \frac{f_\text{no-nodule}}{f_\text{nodule}} & \text{if $c$ is nodule}
\\ 1 & \text{otherwise}
\end{array}\right. \nonumber
\end{align}

In the malignancy detector, we slightly alter the loss function and generalise it to allowing balancing of multiple classes as in Equation~\ref{eq:norm_loss}. We did not have time to retrain the initial nodule detector with this generalised loss, but expect similar performance.

\begin{align}
\textsc{Loss}(p, q) &= -\frac{1}{|C|} \sum_{c \in C} \frac{1}{f_{c}} \cdot p(c) \log q(c) \label{eq:norm_loss}
\\ \text{where}~ & p ~\text{is the predicted distribution,}  \nonumber
\\ & q ~\text{is the true distribution,} \nonumber
\\ & C ~\text{is} ~\{\text{`malignant', `benign', `no-nodule'}\}\text{, and--} \nonumber
\\ & f_c ~\text{is the frequency of class $c$ in the mini-batch.} \nonumber
\end{align}

\subsubsection{Nodule Classifier}

\label{sec:structclass}
Every grid cell reported to contain a nodule by the nodule detector is extracted from the original volume. Note that the nodules classifier works on the detector output and not on the malignancy detector one. This allows the framework to be more robust by ensembling results from the separate networks.

Contiguous grid cells with nodules are assumed to contain parts of the same nodule, and are stitched together. All nodules are scaled and zero-padded to fit into a \cubesize{32} volume.

The nodule classifier takes as input the \cubesize{32} volume with the nodule and classifies it as malignant or benign. As the size of the input is smaller, training the classifier on the detected nodules is significantly easier than on the entire scan volume. The classifier is based on the pre-activation version of ResNet-18 and is described in Table~\ref{resnet18}.

\begin{table}[h]
\centering
\begin{tabular}{lccr}
\toprule
\textbf{Layers} & \textbf{Size, Filters} & \textbf{Stride} & \textbf{Output Size}
\\ \midrule
Convolution 3D & \(7 \times 7 \times 7\), 64 & 2 & \(16 \times 16 \times 16\)
\\
Max Pool 3D & \(3 \times 3 \times 3\) & 2 & \(8 \times 8 \times 8\)
\\ \midrule \addlinespace[0.5em]
Block 1 & \(\begin{bmatrix}
3 \times 3 \times 3, 64 \\
3 \times 3 \times 3, 64 \\
\end{bmatrix} \times 2\) & 1 & \(8 \times 8 \times 8\)
\\ \addlinespace[0.5em]
Block 2 & \(\begin{bmatrix}
3 \times 3 \times 3, 128 \\
3 \times 3 \times 3, 128 \\
\end{bmatrix} \times 2\) & 2 & \(4 \times 4 \times 4\)
\\ \addlinespace[0.5em]
Block 3 & \(\begin{bmatrix}
3 \times 3 \times 3, 256 \\
3 \times 3 \times 3, 256 \\
\end{bmatrix} \times 2\) & 1 & \(4 \times 4 \times 4\)
\\ \addlinespace[0.5em]
Block 4 & \(\begin{bmatrix}
3 \times 3 \times 3, 512 \\
3 \times 3 \times 3, 512 \\
\end{bmatrix} \times 2\) & 1 & \(4 \times 4 \times 4\)
\\ \addlinespace[0.5em] \midrule
Average Pool 3D & Global &  & \(1 \times 1 \times 1\)
\\
Fully Connected & 2 &  & 1
\\
Softmax &  &  & 1
\\ \bottomrule
\end{tabular}
\caption{ResNet-18, modified for nodule classification. The strides are lighter than in the original structure, and we use average pooling at the end.}
\label{resnet18}
\end{table}

\subsubsection{Patient Classifier}
\label{subsubsec:patient_classifier_arch}

To create a single feature vector for each patient, the global patient features from the malignancy nodule detector are combined with the local nodule features from the nodule classifier. 

We aggregate the output of the malignancy detector by combining crops for each patient and constructing a density histogram over the softmax output, with 32 bins for each class in \{`malignant', `benign', `no-nodule'\}. The number of crops containing a nodule is also appended to this feature. We pool the nodule classifier outputs by computing the number of nodules, minimum, maximum, mean, standard deviation, and sum of the softmax output, as well as a 10 bin density histogram. In the case of a patient without any detected nodule, all these values are set to zero.

The features from both networks are then weighted and concatenated into a 113 dimensional vector. This acts as input to a simple neural network with two hidden layers followed by ReLU non-linearity as shown in Table~\ref{tab:patient_classifier}. This network produces the two-class probability of the cancer status of the patient.

The use of histograms to perform pooling in neural networks is a long-accepted technique, dating to the earliest development of neural networks\cite{bishop1995neural}.

\begin{table}[h]
\centering
\begin{tabular}{lrr}
\toprule
\textbf{Layers} & \textbf{Features} & \textbf{Dropout}
\\ \midrule
Fully Connected & 1024 & 0.5
\\ \midrule
Fully Connected & 1024 & 0.5
\\ \midrule
Fully Connected & 2 
\\
Softmax & 
\\ \bottomrule
\end{tabular}
\caption{Patient Classifier with two hidden layers pooling features from the malignancy detector and nodule classifier into a two-class probability of the cancer status of the patient.}
\label{tab:patient_classifier}
\end{table}

\subsection{Training}
\subsubsection{Nodule Detector, LUNA16}
\label{sec:training_nodule_detector}

Each volume crop in LUNA16 is preprocessed then divided into a uniform grid, with each cell of size \cubesize{16}. If the bounding box of a nodule intersects with a grid cell, that cell is deemed to be labelled `has-nodule'; other cells are labelled `no-nodule'. To save time, we sample only 128 random crops from each patient for training, duplicating crops with nodules to maintain class balance.

We train for 100~000 iterations of 24 mini-batches, with a learning rate of 0.01 and weight decay of $10^{-4}$. The Adam optimizer \cite{kingma2014adam} is used with default parameters of $(\beta_1 = 0.9, \beta_2 = 0.999)$.

\subsubsection{Malignancy Detector, KDSB17}

This network is initialized with the weights of the trained nodule detector network, modified to return a distribution over three classes $C = \{$`malignant', `benign', `no-nodule'$\}$. We fine-tune this network on the KDSB17 dataset: cells classified by the nodule detector as `has-nodule' are classified as `malignant' or `benign' depending on the cancer status of the patient; other cells are classified as `no-nodule'. We only train and test on crops that contain a nodule and maintain class balance by duplication. This is an optional step done to save time. We fine-tune this model, first for 20~000 iterations with a learning rate of 0.01, then for 30~000 iterations with a learning rate of 0.001.

\subsubsection{Nodule Classifier}
Over the KDSB17 dataset, we detect between 0 and 10 nodule grid cells per scan. We stack and average detection results from over-lapping crops and consider detections with a confidence above 0.5 as indicating the presence of a nodule. We then extract all the detected nodules from all the patients, and scale with zero-padding to a fixed value of \cubesize{32} for training. Each of these nodules are classified independently. Generally, in a patient with cancer, only a few of the nodules present are malignant. All nodules in patients without cancer are benign. 

\begin{figure}[h]
\includegraphics[width=2cm, height=2cm]{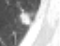}
\includegraphics[width=2cm, height=2cm]{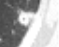}
\includegraphics[width=2cm, height=2cm]{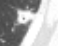}
\includegraphics[width=2cm, height=2cm]{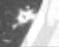}\\
\includegraphics[width=2cm, height=2cm]{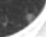}
\includegraphics[width=2cm, height=2cm]{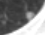}
\includegraphics[width=2cm, height=2cm]{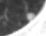}
\includegraphics[width=2cm, height=2cm]{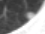}
\centering
\caption{First row: nodule found in a healthy patient. \\
Second row: largest nodule found in a patient with cancer. Size itself cannot distinguish them from healthy ones.\\
The white zones on the right correspond to lung borders.\\}
\label{fig:nodules}
\end{figure}

Benign and malignant nodules are difficult to distinguish, even for experienced radiologists; Figure~\ref{fig:nodules} compares sample cross-sections. Doctors often use the nodule size as a first criteria in nodules examination: cancerous nodules tend to be the largest, and usually larger than a particular threshold \cite{babar}.

For the classifier, we explore different methods of labelling nodules and build the set of malignant nodules using the heuristic measure described in Section~\ref{sec:three_labelling_strategies}.

Malignant nodules are far less prevalent than benign (around 1 for 7). To rebalance the classes, we augment the set of malignant nodules by flipping and 90-degree rotations.

We trained from scratch for 6~000 iterations (stopping early) using the Adam optimizer with a batch size of 32, a learning rate of 0.001, and a weight-decay of $10^{-4}$. We split the training data into an actual training set and a validation one, the latter corresponding to 10\% of original training patients.

\subsubsection{Patient Classifier}

The patient classifier is trained to associate the pooled outputs from the malignancy detector and classifier with the cancer status of the patient, as provided in the KDSB17 dataset. We initialise weights as in \cite{he2015delving}, and train with the Adam optimizer for 2~000 iterations using a learning rate of 0.001, with all the data used as a single batch. To prevent overfitting, we train the patient classifier on an augmented version of the KDSB17 training set (augmented via volume transpose), and use a weight decay of $10^{-4}$. Output is clipped to [0.1, 0.9].

\subsubsection{Strategies for Labelling Training Data}
\label{sec:three_labelling_strategies}

To classify individual nodules we need to obtain labels for each nodule. We do not have any such data, and obtaining radiologist annotation on individual nodules was not feasible. We know that all nodules in patients without cancer are benign; and we use heuristic methods to label nodules within patients with cancer as benign or malignant:

We compare two different heuristic methods to assign labels to nodules: the patient-label strategy, and the largest-nodule strategy. The patient-label strategy is the simplest possible, where we label all nodules from a patient with cancer as malignant. The largest-nodule strategy assumes that in patients with cancer, the largest nodule and all nodules at least some proportion $w$ of that nodule are malignant. We used the latter in the Kaggle competition.

In contrast, the malignancy detector uses the patient-label heuristic. This is a deliberate simplification to avoid the computational overhead of backpropagation through the nodule extractor. The classifier does not incur such an overhead because it operates on nodules after the extractor has assembled them from grid cells.

\section{Results and Discussion}

In medical diagnostics, it is common to present classifier performance using sensitivity (the true positive rate) and specificity (the true negative rate) instead of accuracy. To assess overall classification relevance, we also compute the F1-score. In the KDSB 2017, candidates were evaluated using the log-loss metric. We evaluate each component in the entire pipeline and present our results. 

\subsection{Performance}

\begin{table}[h]
\centering
\begin{tabularx}{\columnwidth}{c X c}
\toprule
\textbf{Rank} & \textbf{Team Name} & \textbf{Log-Loss}
\\ \midrule
   1 & grt123 & 0.39975
\\ 2 & Julian de Wit \& Daniel Hammack & 0.40117
\\ 3 & Aidence & 0.40127
\\ \textbf{41} & \textbf{AIDA (Our Team)} & \textbf{0.52712} 
\\ 50 & Excelsior & 0.55041
\\ 100 & rnrq & 0.60490
\\ 200 & Byeong-wooJeon & 0.61946
\\ \midrule
   & Uniform (0.5) Benchmark & 0.69315
\\ \bottomrule
\end{tabularx}
\caption{Ranking Table, 1972 teams in total.}
\label{tab:results}
\end{table}

We evaluate the patient classifier directly on the stage 1 and stage 2 test sets of the KDSB17 dataset. On stage 1 test data, we observe sensitivity of \textbf{0.719}, specificity of \textbf{0.716}, and Log-Loss of \textbf{0.47707}, ranking our entry as $71^\text{st}$ during the first round of the competition. Sensitivity and specificity are computed by setting the probability threshold separating the two classes on the classifier's output at 0.25.

When testing on stage 2 test data in the second phase of the KDSB17 contest, we observe a Log-Los of \textbf{0.52712} ranking our entry as $41^\text{st}$ out of 1972 teams, placing us in the top $3\%$. As the competition organizers only reported the log loss, we are unable to make a direct comparison of approaches. We present our ranking in Table~\ref{tab:results}. 

During the competition, only 4 features were used from the nodule classifier, number of nodules, mean, std, and sum of the softmax output. Post-competition, we use additional features as described in Section~\ref{subsubsec:patient_classifier_arch}. We present updated results in Table~\ref{tab:componenet_compare} and compare the overall performance of each component and their contribution to the final result.

\begin{table}[h]
\begin{tabularx}{\columnwidth}{ccccc}\toprule & \textbf{Sensitivity} & \textbf{Specificity} & \textbf{F1} & \textbf{Log-Loss}
\\
\midrule
MD + NC & 0.667 & \textbf{0.773} & \textbf{0.598} & \textbf{0.475}
\\
MD & \textbf{0.719} & 0.653 & 0.558 & 0.484
\\  
NC & 0.632 & 0.582 & 0.474 & 0.578
\\
\bottomrule
\end{tabularx}
\caption{Comparing updated performance of the Malignancy Detector (MD), the Nodule Classifier (NC), and their contribution to the final results when pooling through the patient classifier and evaluating on the KDSB17 stage 1 test set. A threshold of 0.25 on the output was used to seperate positive and negative classes.}
\label{tab:componenet_compare}
\end{table}

\subsection{Component Training}
\subsubsection{Nodule \& Malignancy Detector}

To verify the nodule detector's performance, we evaluate on the validation set of LUNA16 and observe sensitivity of \textbf{0.697}, specificity of \textbf{0.999}, and F1-score of \textbf{0.740}. We similarly evaluate the malignancy detector on the stage 1 test set of KDSB17 and observe sensitivity of \textbf{0.317}, specificity of \textbf{0.997}, and F1-score of \textbf{0.269}. The metrics for the malignancy detector are calculated with only the malignant nodules as the positive class.

While the nodule detector performs well, the malignancy detector has comparatively poor performance. This is likely due to the additional class increasing the complexity of the task.

Additionally, the malignancy detector is trained on a version of the KDSB17 dataset where nodules are labelled using a naive method of labelling all nodules in a cancer patient as malignant. This naive patient-labelling method might introduce noise into the groundtruth labels of the annotated dataset, thus impeding the learning of the network. As labelling of nodules is done by the nodule detector, errors in labelling also propagate down to the malignancy detector as well as nodule classifier.

\subsubsection{Nodule Classifier}
We evaluate the classifier's performance using sensitivity, specificity, as well as F1-score.

Initially training with 10~000 steps resulted in almost perfect scores while evaluating the model on training data, indicating a high chance of over-fitting. Additionally, the F1-score started to deteriorate rapidly. To prevent this, we stop training early, before performance on the training set flattens. Table \ref{tab:classifier_metrics_on_testing_set} shows the classifier performance on the testing set after several training durations.

\begin{table}[h]
\begin{tabularx}{\columnwidth}{cccc}\toprule
	 & \textbf{Sensitivity} & \textbf{Specificity} &  \textbf{F1}\\
	\midrule
	4k training steps & 0.385 & \textbf{0.830} & \textbf{0.347}\\
	\emph{6k training steps} & 0.538 & 0.638 & 0.33\\	
	10k training steps & \textbf{0.738} & 0.368 & 0.306\\\bottomrule
\end{tabularx}
\caption{Nodules classification results after different training lengths. These results are obtained on the stage 1 test set. In each case, the best threshold to separate the positive class from the negative one is found on the validation set, and equals to 0.5, 0.45, and 0.8 respectively.}
\label{tab:classifier_metrics_on_testing_set}
\end{table}

For the final architecture, we use a model trained for 6~000 steps. It provides the best trade-off between all three criteria. 

The quality of these classification results directly relies on the quality of the nodules labelling strategy that was used. A strategy that mislabelled many nodules would result in lower classifier performance.

In Table~\ref{tab:labeling_methods} we present the classification results of the patient-label strategy and the largest-nodule strategy (see Section~\ref{sec:three_labelling_strategies}.
)  with $w=90\%$ and $w=70\%$ when applied to the classifier, trained for 6~000 steps.

\begin{table}[h]
\begin{tabularx}{\columnwidth}{cccc}\toprule
	 & \textbf{Sensitivity} & \textbf{Specificity} &  \textbf{F1}\\
	\midrule
	Patient-label & \textbf{0.762} & 0.180 & \textbf{0.434}\\
	Largest ($w=90\%$) & 0.250 & \textbf{0.856} & 0.254\\	
	\emph{Largest ($w=70\%$)}& 0.538 & 0.648 & 0.33\\\bottomrule
\end{tabularx}
\caption{Nodules classification results on stage 1 test set with different ways of labeling nodules. Refer to part 2.3.2 for the explanations related to these different methods. In each case, the best threshold to separate the positive class from the negative one is found on the validation set, and equals to 0.5, 0.55, and 0.45 respectively.}
\label{tab:labeling_methods}
\end{table}

Largest nodule with $w=70\%$ gives an average F1-score and the best trade-off between specificity and sensitivity. This suggests it is the method leading to the least number of mislabelled training nodules. Cancerous nodules are usually the largest ones, and not all nodules are systematically malignant.  In our pipeline, we thus use the largest-nodule strategy $(w=70\%)$ for the classifier.

Changing the strides in the ResNet architecture is essential. With the original stride values from the ResNet-18 architecture, training is less efficient. Convolutional filters become larger than the convolutional feature maps, harming learning and leading to very poor sensitivity and F1-score. In consequence, strides are removed in blocks 1, 2 and 4. Table \ref{tab:classifier_metrics_strides} shows the classifier performance on the testing set after 6~000 training steps.

\begin{table}[h]
\begin{tabularx}{\columnwidth}{ccccc}\toprule
	 & \textbf{Sensitivity} & \textbf{Specificity} &  \textbf{F1}\\
	\midrule
	Original ResNet & 0.169 & \textbf{0.858} & 0.181\\
	\emph{Our architecture} & \textbf{0.538} & 0.648 & \textbf{0.33}\\\bottomrule
\end{tabularx}
\caption{Nodules classification results obtained on the stage 1 test set. We compare results between an original ResNet architecture and the strides modification that we applied. In both cases, the best threshold to separate the positive class from the negative one is found on the validation set, and equals to 0.5 and 0.45 respectively.}
\label{tab:classifier_metrics_strides}
\end{table}

\section{Conclusions and Future Work}
Detecting lung cancer in a full 3D CT-Scan is a challenging task. Directly training a single-stage network is futile, but factoring our solution into multiple stages makes training tractable. Due to imperfect datasets, our approach leveraged the LUNA16 dataset to train a nodule detector, and then refined that detector with the KDSB17 dataset to provide global features. We use that, and pool local features from a separate nodule classifier, to detect lung cancer with high accuracy.

The quality of our method was validated by the competition, in which we placed $41^\text{st}$ out of 1972 teams (top 3\%).

\subsection{Improvements}

There are many ways in which we can extend our method:

Our method makes little use of the peculiarities of cancer nodules, and so will likely improve with advice from medical professionals. \citet{juliandewit}, who placed second in the competition, used 17 different CNNs to extract the diameter, lobulation, spiculation, calcification, sphericity, and other features of nodules. These features are commonly used to classify nodules \cite{armato2011lung}, and help the network better learn about malignancy and cancers.

When fine-tuning the nodule classifier, we rely on heuristic methods to determine which nodules are malignant and which are benign. These heuristic methods have not been experimentally validated, and so remain of dubious quality. We can instead apply unsupervised learning techniques together with a small set of radiologist-labelled nodules to directly learn the difference between malignant and benign nodules.

\subsection{Future Work}

The generality of our method also suggests that it can be adapted to other tumour- and cancer diagnosis problems. The design of our pipeline does not rely on any particular feature of lungs or lung cancer, and so we can easily adapt our pipeline to other nodular cancers, or perhaps other diseases.

We have many more negative examples than positive examples, and so we need to balance our classes to improve classification performance. Current balancing techniques rely on classical data augmentation (flipping, rotation, etc.), though we would like to investigate advanced techniques such as 3D Generative Adversarial Networks (3D GANs) \cite{3dgan}. GANs are a relatively novel invention, and such a fusion technique may yield insight into both GANs and lung cancer. 

Radiologists do not arrive at a diagnosis of lung cancer from a single CT scan \cite{babar}. They diagnose a particular type of lung cancer using a sequence of CT scans over a few months. They match the behaviour of the nodules over time with a particular subtype of lung cancer. To match radiologist-level accuracy on the task, we need to develop a time-varying model of lung cancer that can effectively include a progression of CT scans.

Patient-level priors also significantly affect diagnoses. Age, sex, smoking behaviour, familial co-occurrence, occupation, etc. are all factors that influence the likelihood of developing lung cancer. Adding this heterogeneous data will yield significant improvement in detection performance.

From a clinical perspective, detecting the subtype and optimizing treatment options is a vitally important problem that machine learning can tackle. Our long-term goal is to model the disease itself, so that we can detect it, predict its behaviour, and treat it optimally.

\bibliographystyle{ACM-Reference-Format}
\bibliography{cikm-lung-cancer}


\begin{thebibliography}{00}


\ifx \showCODEN    \undefined \def \showCODEN     #1{\unskip}     \fi
\ifx \showDOI      \undefined \def \showDOI       #1{#1}\fi
\ifx \showISBNx    \undefined \def \showISBNx     #1{\unskip}     \fi
\ifx \showISBNxiii \undefined \def \showISBNxiii  #1{\unskip}     \fi
\ifx \showISSN     \undefined \def \showISSN      #1{\unskip}     \fi
\ifx \showLCCN     \undefined \def \showLCCN      #1{\unskip}     \fi
\ifx \shownote     \undefined \def \shownote      #1{#1}          \fi
\ifx \showarticletitle \undefined \def \showarticletitle #1{#1}   \fi
\ifx \showURL      \undefined \def \showURL       {\relax}        \fi
\providecommand\bibfield[2]{#2}
\providecommand\bibinfo[2]{#2}
\providecommand\natexlab[1]{#1}
\providecommand\showeprint[2][]{arXiv:#2}

\bibitem[\protect\citeauthoryear{??}{Cam}{2016}]%
        {Camelyon16}
 \bibinfo{year}{2016}\natexlab{}.
\newblock \bibinfo{title}{{Camelyon16} Challenge on cancer metastases detection
  in lymph node}.
\newblock   (\bibinfo{year}{2016}).
\newblock
\showURL{%
\url{http://camelyon16.grand-challenge.org}}


\bibitem[\protect\citeauthoryear{??}{nod}{2016}]%
        {noduleinfo}
 \bibinfo{year}{2016}\natexlab{}.
\newblock \bibinfo{title}{Pulmonary nodules}.
\newblock   (\bibinfo{year}{2016}).
\newblock
\showURL{%
\url{https://my.clevelandclinic.org/health/articles/pulmonary-nodules}}


\bibitem[\protect\citeauthoryear{??}{can}{2017}]%
        {cancerus}
 \bibinfo{year}{2017}\natexlab{}.
\newblock \bibinfo{title}{Lung Cancer Fact Sheet}.
\newblock   (\bibinfo{year}{2017}).
\newblock
\showURL{%
\url{http://www.lung.org/lung-health-and-diseases/lung-disease-lookup/lung-cancer/resource-library/lung-cancer-fact-sheet.html}}


\bibitem[\protect\citeauthoryear{Armato, McLennan, Bidaut, McNitt-Gray, Meyer,
  Reeves, Zhao, Aberle, Henschke, Hoffman, et~al\mbox{.}}{Armato
  et~al\mbox{.}}{2011}]%
        {armato2011lung}
\bibfield{author}{\bibinfo{person}{Samuel~G Armato}, \bibinfo{person}{Geoffrey
  McLennan}, \bibinfo{person}{Luc Bidaut}, \bibinfo{person}{Michael~F
  McNitt-Gray}, \bibinfo{person}{Charles~R Meyer}, \bibinfo{person}{Anthony~P
  Reeves}, \bibinfo{person}{Binsheng Zhao}, \bibinfo{person}{Denise~R Aberle},
  \bibinfo{person}{Claudia~I Henschke}, \bibinfo{person}{Eric~A Hoffman}, {and}
  \bibinfo{person}{others}.} \bibinfo{year}{2011}\natexlab{}.
\newblock \showarticletitle{The lung image database consortium (LIDC) and image
  database resource initiative (IDRI): a completed reference database of lung
  nodules on CT scans}.
\newblock \bibinfo{journal}{{\em Medical physics\/}} \bibinfo{volume}{38},
  \bibinfo{number}{2} (\bibinfo{year}{2011}), \bibinfo{pages}{915--931}.
\newblock


\bibitem[\protect\citeauthoryear{Bishop}{Bishop}{1995}]%
        {bishop1995neural}
\bibfield{author}{\bibinfo{person}{Christopher~M Bishop}.}
  \bibinfo{year}{1995}\natexlab{}.
\newblock \bibinfo{booktitle}{{\em Neural networks for pattern recognition}}.
\newblock \bibinfo{publisher}{Oxford university press}.
\newblock


\bibitem[\protect\citeauthoryear{{\c{C}}i{\c{c}}ek, Abdulkadir, Lienkamp, Brox,
  and Ronneberger}{{\c{C}}i{\c{c}}ek et~al\mbox{.}}{2016}]%
        {unet3d}
\bibfield{author}{\bibinfo{person}{{\"O}zg{\"u}n {\c{C}}i{\c{c}}ek},
  \bibinfo{person}{Ahmed Abdulkadir}, \bibinfo{person}{Soeren~S Lienkamp},
  \bibinfo{person}{Thomas Brox}, {and} \bibinfo{person}{Olaf Ronneberger}.}
  \bibinfo{year}{2016}\natexlab{}.
\newblock \showarticletitle{3d u-net: learning dense volumetric segmentation
  from sparse annotation}. In \bibinfo{booktitle}{{\em International Conference
  on Medical Image Computing and Computer-Assisted Intervention}}. Springer,
  \bibinfo{pages}{424--432}.
\newblock


\bibitem[\protect\citeauthoryear{de~Wit}{de~Wit}{2017}]%
        {juliandewit}
\bibfield{author}{\bibinfo{person}{Julian de Wit}.}
  \bibinfo{year}{2017}\natexlab{}.
\newblock \bibinfo{title}{2nd place solution for the 2017 national datascience
  bowl}.
\newblock   (\bibinfo{year}{2017}).
\newblock
\showURL{%
\url{https://juliandewit.github.io/kaggle-ndsb2017/}}


\bibitem[\protect\citeauthoryear{Demir and Yener}{Demir and Yener}{2005}]%
        {histosurvey}
\bibfield{author}{\bibinfo{person}{Cigdem Demir} {and}
  \bibinfo{person}{B{\"u}lent Yener}.} \bibinfo{year}{2005}\natexlab{}.
\newblock \showarticletitle{Automated cancer diagnosis based on
  histopathological images: a systematic survey}.
\newblock \bibinfo{journal}{{\em Rensselaer Polytechnic Institute, Tech.
  Rep\/}} (\bibinfo{year}{2005}).
\newblock


\bibitem[\protect\citeauthoryear{Dou, Chen, Yu, Qin, and Heng}{Dou
  et~al\mbox{.}}{2016}]%
        {fprnodule}
\bibfield{author}{\bibinfo{person}{Qi Dou}, \bibinfo{person}{Hao Chen},
  \bibinfo{person}{Lequan Yu}, \bibinfo{person}{Jing Qin}, {and}
  \bibinfo{person}{Pheng~Ann Heng}.} \bibinfo{year}{2016}\natexlab{}.
\newblock \showarticletitle{Multi-level contextual 3D CNNs for false positive
  reduction in pulmonary nodule detection}.
\newblock \bibinfo{journal}{{\em IEEE Transactions on Biomedical
  Engineering\/}} (\bibinfo{year}{2016}).
\newblock


\bibitem[\protect\citeauthoryear{EF, Jr, P, C, and et~al}{EF
  et~al\mbox{.}}{2014}]%
        {doi:10.1001/jamainternmed.2013.12738}
\bibfield{author}{\bibinfo{person}{Patz EF}, \bibinfo{person}{Jr},
  \bibinfo{person}{Pinsky P}, \bibinfo{person}{Gatsonis C}, {and}
  \bibinfo{person}{et al}.} \bibinfo{year}{2014}\natexlab{}.
\newblock \showarticletitle{Overdiagnosis in low-dose computed tomography
  screening for lung cancer}.
\newblock \bibinfo{journal}{{\em JAMA Internal Medicine\/}}
  \bibinfo{volume}{174}, \bibinfo{number}{2} (\bibinfo{year}{2014}),
  \bibinfo{pages}{269--274}.
\newblock
\showDOI{%
\url{https://doi.org/10.1001/jamainternmed.2013.12738}}
\showeprint{/data/journals/intemed/929736/ioi130125.pdf}


\bibitem[\protect\citeauthoryear{grt123}{grt123}{2017}]%
        {grt123}
\bibfield{author}{\bibinfo{person}{grt123}.} \bibinfo{year}{2017}\natexlab{}.
\newblock \bibinfo{title}{Solution of the ’grt123’ Team}.
\newblock   (\bibinfo{year}{2017}).
\newblock
\showURL{%
\url{https://github.com/lfz/DSB2017/blob/master/solution-grt123-team.pdf}}


\bibitem[\protect\citeauthoryear{Hao, Qi, Lequan, Pheng-Ann, Jason~D., Lily,
  and Martin~C.}{Hao et~al\mbox{.}}{2016}]%
        {voxresnet}
\bibfield{author}{\bibinfo{person}{Chen Hao}, \bibinfo{person}{Dou Qi},
  \bibinfo{person}{Yu Lequan}, \bibinfo{person}{Hengorrado Pheng-Ann},
  \bibinfo{person}{Hipp Jason~D.}, \bibinfo{person}{Peng Lily}, {and}
  \bibinfo{person}{Stumpe Martin~C.}} \bibinfo{year}{2016}\natexlab{}.
\newblock \showarticletitle{VoxResNet: Deep Voxelwise Residual Networks for
  Volumetric Brain Segmentation}.
\newblock  (\bibinfo{year}{2016}).
\newblock
\showDOI{%
\url{https://doi.org/abs/1608.05895}}


\bibitem[\protect\citeauthoryear{He, Zhang, Ren, and Sun}{He
  et~al\mbox{.}}{2015}]%
        {he2015delving}
\bibfield{author}{\bibinfo{person}{Kaiming He}, \bibinfo{person}{Xiangyu
  Zhang}, \bibinfo{person}{Shaoqing Ren}, {and} \bibinfo{person}{Jian Sun}.}
  \bibinfo{year}{2015}\natexlab{}.
\newblock \showarticletitle{Delving deep into rectifiers: Surpassing
  human-level performance on imagenet classification}. In
  \bibinfo{booktitle}{{\em Proceedings of the IEEE international conference on
  computer vision}}. \bibinfo{pages}{1026--1034}.
\newblock


\bibitem[\protect\citeauthoryear{He, Zhang, Ren, and Sun}{He
  et~al\mbox{.}}{2016a}]%
        {he2016deep}
\bibfield{author}{\bibinfo{person}{Kaiming He}, \bibinfo{person}{Xiangyu
  Zhang}, \bibinfo{person}{Shaoqing Ren}, {and} \bibinfo{person}{Jian Sun}.}
  \bibinfo{year}{2016}\natexlab{a}.
\newblock \showarticletitle{Deep residual learning for image recognition}. In
  \bibinfo{booktitle}{{\em Proceedings of the IEEE Conference on Computer
  Vision and Pattern Recognition}}. \bibinfo{pages}{770--778}.
\newblock


\bibitem[\protect\citeauthoryear{He, Zhang, Ren, and Sun}{He
  et~al\mbox{.}}{2016b}]%
        {he2016identity}
\bibfield{author}{\bibinfo{person}{Kaiming He}, \bibinfo{person}{Xiangyu
  Zhang}, \bibinfo{person}{Shaoqing Ren}, {and} \bibinfo{person}{Jian Sun}.}
  \bibinfo{year}{2016}\natexlab{b}.
\newblock \showarticletitle{Identity mappings in deep residual networks}. In
  \bibinfo{booktitle}{{\em European Conference on Computer Vision}}. Springer,
  \bibinfo{pages}{630--645}.
\newblock


\bibitem[\protect\citeauthoryear{Kaggle}{Kaggle}{2017}]%
        {kaggle_2017}
\bibfield{author}{\bibinfo{person}{Kaggle}.} \bibinfo{year}{2017}\natexlab{}.
\newblock \bibinfo{title}{Kaggle Data Science Bowl 2017}.
\newblock   (\bibinfo{year}{2017}).
\newblock
\showURL{%
\url{https://www.kaggle.com/c/data-science-bowl-2017}}
\newblock
\shownote{[Accessed 26-April-2017].}


\bibitem[\protect\citeauthoryear{Kayalibay, Jensen, and van~der
  Smagt}{Kayalibay et~al\mbox{.}}{2017}]%
        {KayalibayJS17}
\bibfield{author}{\bibinfo{person}{Baris Kayalibay}, \bibinfo{person}{Grady
  Jensen}, {and} \bibinfo{person}{Patrick van~der Smagt}.}
  \bibinfo{year}{2017}\natexlab{}.
\newblock \showarticletitle{CNN-based Segmentation of Medical Imaging Data}.
\newblock \bibinfo{journal}{{\em CoRR\/}}  \bibinfo{volume}{abs/1701.03056}
  (\bibinfo{year}{2017}).
\newblock
\showURL{%
\url{http://arxiv.org/abs/1701.03056}}


\bibitem[\protect\citeauthoryear{Khalvati, Wong, and Haider}{Khalvati
  et~al\mbox{.}}{2015}]%
        {prostate}
\bibfield{author}{\bibinfo{person}{Farzad Khalvati}, \bibinfo{person}{Alexander
  Wong}, {and} \bibinfo{person}{Masoom~A Haider}.}
  \bibinfo{year}{2015}\natexlab{}.
\newblock \showarticletitle{Automated prostate cancer detection via
  comprehensive multi-parametric magnetic resonance imaging texture feature
  models}.
\newblock \bibinfo{journal}{{\em BMC medical imaging\/}} \bibinfo{volume}{15},
  \bibinfo{number}{1} (\bibinfo{year}{2015}), \bibinfo{pages}{27}.
\newblock


\bibitem[\protect\citeauthoryear{Kingma and Ba}{Kingma and Ba}{2014}]%
        {kingma2014adam}
\bibfield{author}{\bibinfo{person}{Diederik Kingma} {and}
  \bibinfo{person}{Jimmy Ba}.} \bibinfo{year}{2014}\natexlab{}.
\newblock \showarticletitle{Adam: A method for stochastic optimization}.
\newblock \bibinfo{journal}{{\em arXiv preprint arXiv:1412.6980\/}}
  (\bibinfo{year}{2014}).
\newblock


\bibitem[\protect\citeauthoryear{Krizhevsky, Sutskever, and Hinton}{Krizhevsky
  et~al\mbox{.}}{2012}]%
        {NIPS2012_4824}
\bibfield{author}{\bibinfo{person}{Alex Krizhevsky}, \bibinfo{person}{Ilya
  Sutskever}, {and} \bibinfo{person}{Geoffrey~E Hinton}.}
  \bibinfo{year}{2012}\natexlab{}.
\newblock \showarticletitle{ImageNet Classification with Deep Convolutional
  Neural Networks}.
\newblock In \bibinfo{booktitle}{{\em Advances in Neural Information Processing
  Systems 25}}, \bibfield{editor}{\bibinfo{person}{F.~Pereira},
  \bibinfo{person}{C.~J.~C. Burges}, \bibinfo{person}{L.~Bottou}, {and}
  \bibinfo{person}{K.~Q. Weinberger}} (Eds.). \bibinfo{publisher}{Curran
  Associates, Inc.}, \bibinfo{pages}{1097--1105}.
\newblock
\showURL{%
\url{http://papers.nips.cc/paper/4824-imagenet-classification-with-deep-convolutional-neural-networks.pdf}}


\bibitem[\protect\citeauthoryear{Kumar, Wong, and Clausi}{Kumar
  et~al\mbox{.}}{2015}]%
        {nodulesAE}
\bibfield{author}{\bibinfo{person}{Devinder Kumar}, \bibinfo{person}{Alexander
  Wong}, {and} \bibinfo{person}{David~A Clausi}.}
  \bibinfo{year}{2015}\natexlab{}.
\newblock \showarticletitle{Lung nodule classification using deep features in
  CT images}. In \bibinfo{booktitle}{{\em Computer and Robot Vision (CRV), 2015
  12th Conference on}}. IEEE, \bibinfo{pages}{133--138}.
\newblock


\bibitem[\protect\citeauthoryear{Liao, Gao, Oto, and Shen}{Liao
  et~al\mbox{.}}{2013}]%
        {prostatesegm}
\bibfield{author}{\bibinfo{person}{Shu Liao}, \bibinfo{person}{Yaozong Gao},
  \bibinfo{person}{Aytekin Oto}, {and} \bibinfo{person}{Dinggang Shen}.}
  \bibinfo{year}{2013}\natexlab{}.
\newblock \showarticletitle{Representation learning: a unified deep learning
  framework for automatic prostate MR segmentation}. In
  \bibinfo{booktitle}{{\em International Conference on Medical Image Computing
  and Computer-Assisted Intervention}}. Springer, \bibinfo{pages}{254--261}.
\newblock


\bibitem[\protect\citeauthoryear{Milletari, Navab, and Ahmadi}{Milletari
  et~al\mbox{.}}{2016}]%
        {vnet}
\bibfield{author}{\bibinfo{person}{Fausto Milletari}, \bibinfo{person}{Nassir
  Navab}, {and} \bibinfo{person}{Seyed-Ahmad Ahmadi}.}
  \bibinfo{year}{2016}\natexlab{}.
\newblock \showarticletitle{V-net: Fully convolutional neural networks for
  volumetric medical image segmentation}. In \bibinfo{booktitle}{{\em 3D Vision
  (3DV), 2016 Fourth International Conference on}}. IEEE,
  \bibinfo{pages}{565--571}.
\newblock


\bibitem[\protect\citeauthoryear{Moira, Robbert, Michael, Jeroen, and
  Guido}{Moira et~al\mbox{.}}{2016}]%
        {znet}
\bibfield{author}{\bibinfo{person}{Berens Moira}, \bibinfo{person}{van
  der~Gugten Robbert}, \bibinfo{person}{de~Kaste Michael},
  \bibinfo{person}{Manders Jeroen}, {and} \bibinfo{person}{Zuidhof Guido}.}
  \bibinfo{year}{2016}\natexlab{}.
\newblock \showarticletitle{ZNET - LUNG NODULE DETECTION}.
\newblock  (\bibinfo{year}{2016}).
\newblock


\bibitem[\protect\citeauthoryear{Nazir}{Nazir}{2017}]%
        {babar}
\bibfield{author}{\bibinfo{person}{Babar Nazir}.}
  \bibinfo{year}{2017}\natexlab{}.
\newblock \bibinfo{howpublished}{personal communication}.
  (\bibinfo{year}{2017}).
\newblock


\bibitem[\protect\citeauthoryear{Redmon, Divvala, Girshick, and Farhadi}{Redmon
  et~al\mbox{.}}{2016}]%
        {redmon2016you}
\bibfield{author}{\bibinfo{person}{Joseph Redmon}, \bibinfo{person}{Santosh
  Divvala}, \bibinfo{person}{Ross Girshick}, {and} \bibinfo{person}{Ali
  Farhadi}.} \bibinfo{year}{2016}\natexlab{}.
\newblock \showarticletitle{You only look once: Unified, real-time object
  detection}. In \bibinfo{booktitle}{{\em Proceedings of the IEEE Conference on
  Computer Vision and Pattern Recognition}}. \bibinfo{pages}{779--788}.
\newblock


\bibitem[\protect\citeauthoryear{Ren, He, Girshick, and Sun}{Ren
  et~al\mbox{.}}{2015}]%
        {renNIPS15fasterrcnn}
\bibfield{author}{\bibinfo{person}{Shaoqing Ren}, \bibinfo{person}{Kaiming He},
  \bibinfo{person}{Ross Girshick}, {and} \bibinfo{person}{Jian Sun}.}
  \bibinfo{year}{2015}\natexlab{}.
\newblock \showarticletitle{Faster {R-CNN}: Towards Real-Time Object Detection
  with Region Proposal Networks}. In \bibinfo{booktitle}{{\em Advances in
  Neural Information Processing Systems ({NIPS})}}.
\newblock


\bibitem[\protect\citeauthoryear{Ronneberger, Fischer, and Brox}{Ronneberger
  et~al\mbox{.}}{2015}]%
        {unet}
\bibfield{author}{\bibinfo{person}{Olaf Ronneberger}, \bibinfo{person}{Philipp
  Fischer}, {and} \bibinfo{person}{Thomas Brox}.}
  \bibinfo{year}{2015}\natexlab{}.
\newblock \showarticletitle{U-net: Convolutional networks for biomedical image
  segmentation}. In \bibinfo{booktitle}{{\em International Conference on
  Medical Image Computing and Computer-Assisted Intervention}}. Springer,
  \bibinfo{pages}{234--241}.
\newblock


\bibitem[\protect\citeauthoryear{Setio, Traverso, de~Bel, Berens, van~den
  Bogaard, Cerello, Chen, Dou, Fantacci, Geurts, van~der Gugten, Heng, Jansen,
  de~Kaste, Kotov, Lin, Manders, S{\'{o}}nora{-}Mengana,
  Garc{\'{\i}}a{-}Naranjo, Prokop, Saletta, Schaefer{-}Prokop, Scholten,
  Scholten, Snoeren, Torres, Vandemeulebroucke, Walasek, Zuidhof, van Ginneken,
  and Jacobs}{Setio et~al\mbox{.}}{2016}]%
        {DBLP:journals/corr/SetioTBBBC0DFGG16}
\bibfield{author}{\bibinfo{person}{Arnaud Arindra~Adiyoso Setio},
  \bibinfo{person}{Alberto Traverso}, \bibinfo{person}{Thomas de Bel},
  \bibinfo{person}{Moira S.~N. Berens}, \bibinfo{person}{Cas van~den Bogaard},
  \bibinfo{person}{Piergiorgio Cerello}, \bibinfo{person}{Hao Chen},
  \bibinfo{person}{Qi Dou}, \bibinfo{person}{Maria~Evelina Fantacci},
  \bibinfo{person}{Bram Geurts}, \bibinfo{person}{Robbert van~der Gugten},
  \bibinfo{person}{Pheng{-}Ann Heng}, \bibinfo{person}{Bart Jansen},
  \bibinfo{person}{Michael M.~J. de Kaste}, \bibinfo{person}{Valentin Kotov},
  \bibinfo{person}{Jack~Yu{-}Hung Lin}, \bibinfo{person}{Jeroen T. M.~C.
  Manders}, \bibinfo{person}{Alexander S{\'{o}}nora{-}Mengana},
  \bibinfo{person}{Juan~Carlos Garc{\'{\i}}a{-}Naranjo},
  \bibinfo{person}{Mathias Prokop}, \bibinfo{person}{Marco Saletta},
  \bibinfo{person}{Cornelia Schaefer{-}Prokop}, \bibinfo{person}{Ernst~Th.
  Scholten}, \bibinfo{person}{Luuk Scholten}, \bibinfo{person}{Miranda~M.
  Snoeren}, \bibinfo{person}{Ernesto~Lopez Torres}, \bibinfo{person}{Jef
  Vandemeulebroucke}, \bibinfo{person}{Nicole Walasek}, \bibinfo{person}{Guido
  C.~A. Zuidhof}, \bibinfo{person}{Bram van Ginneken}, {and}
  \bibinfo{person}{Colin Jacobs}.} \bibinfo{year}{2016}\natexlab{}.
\newblock \showarticletitle{Validation, comparison, and combination of
  algorithms for automatic detection of pulmonary nodules in computed
  tomography images: the {LUNA16} challenge}.
\newblock \bibinfo{journal}{{\em CoRR\/}}  \bibinfo{volume}{abs/1612.08012}
  (\bibinfo{year}{2016}).
\newblock
\showURL{%
\url{http://arxiv.org/abs/1612.08012}}


\bibitem[\protect\citeauthoryear{Simonyan and Zisserman}{Simonyan and
  Zisserman}{2014}]%
        {Simonyan14c}
\bibfield{author}{\bibinfo{person}{K. Simonyan} {and} \bibinfo{person}{A.
  Zisserman}.} \bibinfo{year}{2014}\natexlab{}.
\newblock \showarticletitle{Very Deep Convolutional Networks for Large-Scale
  Image Recognition}.
\newblock \bibinfo{journal}{{\em CoRR\/}}  \bibinfo{volume}{abs/1409.1556}
  (\bibinfo{year}{2014}).
\newblock


\bibitem[\protect\citeauthoryear{Team}{Team}{2011}]%
        {doi:10.1056/NEJMoa1102873}
\bibfield{author}{\bibinfo{person}{The National Lung Screening Trial~Research
  Team}.} \bibinfo{year}{2011}\natexlab{}.
\newblock \showarticletitle{Reduced Lung-Cancer Mortality with Low-Dose
  Computed Tomographic Screening}.
\newblock \bibinfo{journal}{{\em New England Journal of Medicine\/}}
  \bibinfo{volume}{365}, \bibinfo{number}{5} (\bibinfo{year}{2011}),
  \bibinfo{pages}{395--409}.
\newblock
\showDOI{%
\url{https://doi.org/10.1056/NEJMoa1102873}}
\showeprint{http://dx.doi.org/10.1056/NEJMoa1102873}
\newblock
\shownote{PMID: 21714641.}


\bibitem[\protect\citeauthoryear{Torre, Bray, Siegel, Ferlay, Lortet-Tieulent,
  and Jemal}{Torre et~al\mbox{.}}{2015}]%
        {CAAC:CAAC21262}
\bibfield{author}{\bibinfo{person}{Lindsey~A. Torre}, \bibinfo{person}{Freddie
  Bray}, \bibinfo{person}{Rebecca~L. Siegel}, \bibinfo{person}{Jacques Ferlay},
  \bibinfo{person}{Joannie Lortet-Tieulent}, {and} \bibinfo{person}{Ahmedin
  Jemal}.} \bibinfo{year}{2015}\natexlab{}.
\newblock \showarticletitle{Global cancer statistics, 2012}.
\newblock \bibinfo{journal}{{\em CA: A Cancer Journal for Clinicians\/}}
  \bibinfo{volume}{65}, \bibinfo{number}{2} (\bibinfo{year}{2015}),
  \bibinfo{pages}{87--108}.
\newblock
\showISSN{1542-4863}
\showDOI{%
\url{https://doi.org/10.3322/caac.21262}}


\bibitem[\protect\citeauthoryear{Wu, Zhang, Xue, Freeman, and Tenenbaum}{Wu
  et~al\mbox{.}}{2016}]%
        {3dgan}
\bibfield{author}{\bibinfo{person}{Jiajun Wu}, \bibinfo{person}{Chengkai
  Zhang}, \bibinfo{person}{Tianfan Xue}, \bibinfo{person}{William~T Freeman},
  {and} \bibinfo{person}{Joshua~B Tenenbaum}.} \bibinfo{year}{2016}\natexlab{}.
\newblock \showarticletitle{Learning a probabilistic latent space of object
  shapes via 3d generative-adversarial modeling}. In \bibinfo{booktitle}{{\em
  Advances in Neural Information Processing Systems}}. \bibinfo{pages}{82--90}.
\newblock


\bibitem[\protect\citeauthoryear{Yun, Krishna, Mohammad, George~E., Timo,
  Aleksey, Subhashini, Aleksei, Philip~Q., Greg~S., Jason~D., Lily, and
  Martin~C.}{Yun et~al\mbox{.}}{2017}]%
        {googlemetastasis}
\bibfield{author}{\bibinfo{person}{Liu Yun}, \bibinfo{person}{Gadepalli
  Krishna}, \bibinfo{person}{Norouzi Mohammad}, \bibinfo{person}{Dahl
  George~E.}, \bibinfo{person}{Kohlberger Timo}, \bibinfo{person}{Boyko
  Aleksey}, \bibinfo{person}{Venugopalan Subhashini}, \bibinfo{person}{Timofeev
  Aleksei}, \bibinfo{person}{Nelson Philip~Q.}, \bibinfo{person}{Corrado
  Greg~S.}, \bibinfo{person}{Hipp Jason~D.}, \bibinfo{person}{Peng Lily}, {and}
  \bibinfo{person}{Stumpe Martin~C.}} \bibinfo{year}{2017}\natexlab{}.
\newblock \showarticletitle{Detecting Cancer Metastases on Gigapixel Pathology
  Images}.
\newblock  (\bibinfo{year}{2017}).
\newblock
\showDOI{%
\url{https://doi.org/abs/1703.02442}}


\end{thebibliography}

\end{document}